\documentclass[11pt,a4paper]{article}
\usepackage[hyperref]{naaclhlt2019}
\usepackage{times}
\usepackage{latexsym}
\usepackage{amsmath}
\usepackage{graphicx}
\usepackage{booktabs}
\usepackage{multirow}
\usepackage{enumitem}
\usepackage{colortbl}
\usepackage{xcolor}
\usepackage{url}

\aclfinalcopy % Uncomment this line for the final submission
\newcommand{\FT}[0]{\includegraphics[width=.022\textwidth]{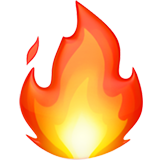}}
\newcommand{\FE}[0]{\includegraphics[width=.022\textwidth]{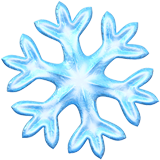}}

\newcommand{\warning}[0]{\includegraphics[width=.022\textwidth]{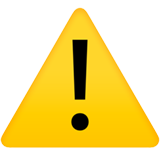}}

\definecolor{fe_blue}{RGB}{152, 236, 249}
\definecolor{fe_red}{RGB}{255, 158, 158}

\title{To Tune or Not to Tune? \\ Adapting Pretrained Representations to Diverse Tasks}

\author{Matthew Peters\textsuperscript{1}\footnotemark, Sebastian Ruder\textsuperscript{2,3}$^\dagger$\footnotemark[1], and Noah A. Smith\textsuperscript{1,4} \\
  \textsuperscript{1}Allen Institute for Artificial Intelligence, Seattle, USA \\
  \textsuperscript{2}Insight Research Centre, National University of Ireland, Galway, Ireland \\
  \textsuperscript{3}Aylien Ltd., Dublin, Ireland \\
  \textsuperscript{4}Paul G.~Allen School of CSE, University of Washington, Seattle, USA \\
  {\tt \{matthewp,noah\}@allenai.org}, {\tt sebastian@ruder.io}
}

\date{}

\begin{document}
\maketitle

\begin{abstract}
%We consider the question of how to best transfer pretrained contextual representations (e.g. ELMo and BERT) to a target task.
While most previous work has focused on different pretraining objectives and architectures for transfer learning, we ask how to best adapt the pretrained model to a given target task. 
We focus on the two most common forms of adaptation, feature extraction (where the pretrained weights are frozen), and directly fine-tuning the pretrained model.
% When transferring pretrained contextual representations (e.g. ELMo and BERT) to a target task, is it better to extract features (by freezing the weights of the pretrained model) or directly fine-tune the pretrained model?
% Seb: We should maybe mention the additional task layers in a second sentence as they are a separate choice from feature extraction / fine-tuning and their relation might not be directly obvious.
%Our empirical results across diverse NLP tasks with two state-of-the-art models show that fine-tuning outperforms feature extraction when the pretraining task and target task are most similar, but both approaches have comparable performance when the target and pretraining tasks are dissimilar.
Our empirical results across diverse NLP tasks with two state-of-the-art models show that the relative performance of fine-tuning vs. feature extraction depends on the similarity of the pretraining and target tasks.
%Our empirical results across diverse NLP tasks with two state-of-the-art models show that in most cases both approaches have comparable performance, except when the pretraining task and target task are most similar.
%\nascomment{do you want to comment that this is a bit surprising/counterintuitive and hint that you'll try to give an explanation?  maybe:  We explore possible explanations for this surprising finding.}
We explore possible explanations for this finding and provide a set of adaptation guidelines for the NLP practitioner.
%We explore possible explanations for this finding and provide a set of practical adaptation guidelines to the NLP practitioner.
% Using diagnostic classifiers to measure the similarity between pre-training and target tasks, we show that language modeling learns generally transferable features, and that the next-sentence prediction task in BERT is particularly well suited for adaptation to paraphrase and similarity tasks.
% Add highlights, take home message
\end{abstract}

\newenvironment{starfootnotes}
  {\par\edef\savedfootnotenumber{\number\value{footnote}}
   \renewcommand{\thefootnote}{$\star$} 
   \setcounter{footnote}{0}}
  {\par\setcounter{footnote}{\savedfootnotenumber}}
  
\begin{starfootnotes}
\footnotetext{The first two authors contributed equally.}
\end{starfootnotes}

\newcommand{\customfootnotetext}[2]{{% Group to localize change to footnote
  \renewcommand{\thefootnote}{#1}% Update footnote counter representation
  \footnotetext[0]{#2}}}% Print footnote text
  
\customfootnotetext{$\dagger$}{Sebastian is now at DeepMind.}

% \FEEDBACK 11/30/2018 start with question and define features vs fine-tune in first sentence
% "diagnostic classifier" is under specified, need to define
% maybe add "surprising" or "counter-intuitive" to the results
% What's the punchline??
% be more explicit with results, e.g. use 3rd sentence to say "In the case of ELMo, a pre-trained language model, both feature and fine-tuning approaches have comparable performance."
% maybe say that fine-tuning with ELMo is hard / very hyperparameter sensitive
% can we give any practical advice in abstract?

%After pre-training, the most relevant information for a given target task is typically held in the internal layers of deep networks.  During fine tuning, this information is effectively moved from the internal layers to the top representation layer for prediction, in a similar manner to how feature based approaches that expose internal network representations move this information upward in the network for prediction.

%There has been a significant amount of prior work regarding different pre-training objetives, but little work focused on how to transfer the pre-trained model to a target task. 

\section{Introduction}

%Sequential inductive transfer learning \cite{Pan2010} consists of two stages: a \emph{pretraining} and an \emph{adaptation} phase. During pretraining, the model accumulates general-purpose information. This information is then transferred to a new task in the adaptation phase.
Sequential inductive transfer learning \cite{Pan2010} consists of two stages: \emph{pretraining}, in which the model learns a general-purpose representation of inputs, and  \emph{adaptation}, in which the representation is  transferred to a new task. 
% PAPERS TO CITE: word2vec, skip-thought, elmo, bert
%Most previous work has focused on different pretraining objectives for learning uncontextual word representations \cite{word2vec}, sentence representations \cite{Kiros2015a}, contextual word representations \cite[ELMo; ][]{Peters2018}, and combinations of sentence and contextual word representations \cite[BERT; ][]{Devlin2018}.
Most previous work in NLP has focused on different pretraining objectives for learning word or sentence representations  \cite{word2vec, Kiros2015a}.
%Much recent work has focused on investigating different pretraining tasks for NLP \cite{Conneau2017,Felbo2017}. 
%These range from supervised tasks, e.g. natural language inference \cite{Conneau2017} 
%and machine translation \cite{Mccann2017},
% to distantly supervised tasks, e.g. emoji prediction \cite{Felbo2017}
%and pivot prediction \cite{Ziser2018}
% to completely unsupervised tasks, such as language modeling \cite{Dai2015a}.

%Pretrained language model representations in particular have achieved remarkable results across a wide range range of NLP tasks \cite{Peters2018,Howard2018}.
 
Few works, however, have focused on the adaptation phase.
%Updating pretrained word embeddings has proven more fruitful than keeping them static \cite{Kim2014}, but it is unclear how to best adapt deep pretrained representations.
There are two main paradigms for adaptation: \emph{feature extraction} and \emph{fine-tuning}.
In feature extraction (\FE) the model's weights are `frozen' and the pretrained representations are used in a downstream model similar to classic feature-based approaches \cite{koehn2003statistical}. %\nascomment{this is an odd choice.  I'd either go with one of the earlier classics by Ratnaparkhi, or phrase-based MT papers where IBM models were trained first, then used to compute features for a MERT-trained linear model; alternately you might be thinking of Charniak and Johnson 2005, but I think that's less apt}.
%\FE~involves `freezing' a model's layers and using the model's hidden representation for each word as feature in a downstream model \cite{Peters2018} similar to classic feature-based approaches \cite{charniak2000maximum}.
Alternatively, a pretrained model's parameters can be unfrozen and fine-tuned (\FT) on a new task \cite{Dai2015SemisupervisedSL}.
%Howard2018,Radford2018,Devlin2018
Both have benefits: \FE~enables use of task-specific model architectures and may be computationally cheaper as features only need to be computed once.
On the other hand, \FT~is convenient as it may allow us to adapt a general-purpose representation to many different tasks.
%\FE~and \FT~are common considerations when using pretrained word embeddings where they correspond to keeping word embeddings static and non-static respectively. 
%On the other hand, \FT~is convenient as it allows us to apply the same model to many tasks. \FE~and \FT~are common considerations when using pretrained word embeddings where they correspond to keeping word embeddings static and non-static respectively. 

\begin{table}[]
\resizebox{\columnwidth}{!}{%
\begin{tabular}{l c c l}
\toprule
\multicolumn{3}{c}{\textbf{Conditions}} & \multicolumn{1}{c}{\multirow{2}{*}{\textbf{Guidelines}}} \\
%\cmidrule{1-3}
Pretrain & Adapt. & Task &  \\
\midrule
Any & \FE & Any & Add many task parameters \\
\midrule
\multirow{2}{*}{Any} & \multirow{2}{*}{\FT} & \multirow{2}{*}{Any} & Add minimal task parameters \\
 &  &  & \warning~Hyper-parameters \\
\midrule
Any & Any & Seq. / clas. & \FE~and~\FT~have similar performance \\
ELMo & Any & Sent. pair & use \FE \\
BERT & Any & Sent. pair & use \FT \\

%Pretrain & Adapt. & Task & Guidelines \\
%\midrule
%Any & \FE & Any & Add many task parameters \\
%\midrule
%\multirow{2}{*}{Any} & \multirow{2}{*}{\FT} & \multirow{2}{*}{Any} & Add minimal task parameters \\
% &  &  & \warning~Hyper-parameters \\
%\midrule
%Any & Any & Seq. / clas. & \FE~and~\FT~have similar performance \\
%ELMo & Any & Sent. pair & use \FE \\
%BERT & Any & Sent. pair & use \FT \\

%\midrule

%\multirow{3}{*}{\begin{tabular}[c]{@{}l@{}}LSTM\\ (e.g. ELMo),\\ default: \FE\end{tabular}} & \multirow{3}{*}{\FT} & \multirow{2}{*}{Seq. / clas.} & ULMFiT techniques, \\
%& & & \warning~Hyper-parameters\\
% &  & Sent. pair & $\uparrow$ + self-attention \\
%\midrule
%\begin{tabular}[c]{@{}l@{}}Transformer\\ (e.g. BERT),\\ default: \FT\end{tabular} & \FT & \question & Learning rate schedule \\
%\midrule
%\question & \FT & \question & Few additional parameters \\
%\midrule
%\question & \FE & \question & Many additional parameters \\
\bottomrule
\end{tabular}%
}
\caption{This paper's guidelines for using feature extraction (\FE) and fine-tuning (\FT) with ELMo and BERT. Seq.: sequence labeling. Clas.: classification. Sent. pair: sentence pair tasks.}
\label{tab:guidelines}
\end{table}

%\FE add many task parameters, \FT add minimal task parameters \\
%\warning~Hyper-parameters when \FT, use triangular learning rate schedule \\
%For sequence tagging, classification, \FE and \FT about same performance \\
%For sentence pair tasks, use \FE w/ ELMo, \FT with BERT \\

% \begin{figure}
% 	\centering
%   	\includegraphics[width=\linewidth]{fig/flowchart_color.png}}
%   	\caption{A flowchart for adapting pretrained representations (preferred choices have \textbf{bolded} arrows). \nascomment{I am not in love with this.  why not just make two standalone flowcharts?}
%   	\label{fig:flowchart}}
% \end{figure}

Gaining a better understanding of the adaptation phase is key in making the most use out of pretrained representations.
To this end, we compare two state-of-the-art pretrained models, ELMo \cite{Peters2018} and BERT \cite{Devlin2018} using both \FE~and \FT~across seven diverse tasks including named entity recognition, natural language inference (NLI), and paraphrase detection.
We seek to characterize the conditions under which one approach substantially outperforms the other, and whether it is dependent on the pretraining objective or target task.
We find that \FE~and \FT~have comparable performance in most cases, except when the source and target tasks are either highly similar or highly dissimilar.
%We find that \FE~and \FT~have comparable performance in most cases, except when the source and target tasks are either highly similar or highly dissimilar.
%We find that \FT~outperforms \FE~when source and target tasks are similar but both perform comparably otherwise.
We furthermore shed light on the practical challenges of adaptation and provide a set of guidelines to the NLP practitioner, as summarized in Table \ref{tab:guidelines}.

\section{Pretraining and Adaptation}

% KEY POINTS:
% WORD vs sentence, most based on distributional hypothesis
% sentence: skip thought/quick thought
% similarity of source vs target task -> quality of transfer
% LM: learns good word representations
% BERT: combinaton
% most transferable features are somewhere in middle layers, so FE should try to surface them

While pretraining tasks have been designed with particular downstream tasks in mind \cite{Felbo2017}, we focus on pretraining tasks that seek to induce \emph{universal} representations suitable for any downstream task.
%Some pretraining tasks have been designed with particular downstream tasks in mind \cite{Yang2017b,Felbo2017}. We focus on pretraining tasks that seek to induce \emph{universal} representations suitable for any downstream task and discuss how they have been used in the adaptation phase.

\paragraph{Word representations} Pretrained word vectors \cite{Turian2010WordRA,Pennington2014GloveGV} have been an essential component in state-of-the-art NLP systems. Word representations are often fixed and fed into a task specific model (\FE), although \FT~can provide improvements \cite{Kim2014}. Recently, contextual word representations learned supervisedly \cite[e.g., through machine translation; ][]{Mccann2017} or unsupervisedly \cite[typically through language modeling;][]{Peters2018} have significantly improved over noncontextual vectors.

%Some pretraining tasks have been designed with particular downstream tasks in mind \cite{Yang2017b,Felbo2017}. We focus on pretraining tasks that seek to induce \emph{universal} representations suitable for any downstream task and discuss how they have been used in the adaptation phase.

\paragraph{Sentence embedding methods} Such methods learn sentence representations via different pretraining objectives such as previous/next sentence prediction \cite{Kiros2015a,Logeswaran2018}, NLI \cite{Conneau2017}, or a combination of objectives \cite{Subramanian2018}. During the adaptation phase, the sentence representation is typically provided as input to a linear classifier (\FE).
LM pretraining with \FT~has also been successfully applied to sentence level tasks.   \citet[][ULMFiT]{Howard2018} propose techniques for fine-tuning a LM, including triangular learning rate schedules and discriminative fine-tuning, which uses lower learning rates for lower layers.
\citet{Radford2018} extend LM-\FT~to additional sentence and sentence-pair tasks.

%\paragraph{Language modeling (LM)} ELMo \cite{Peters2018} pretrains a 2-layer bidirectional LSTM language model (LM). For adaptation, a linear combination of the word representation at each layer is fed to a task-specific model. ULMFiT \cite{Howard2018} propose techniques for fine-tuning a similar LM: the slanted triangular learning rate schedule, which quickly increases and then linearly anneals the learning rate; discriminative fine-tuning, which uses lower learning rates for lower layers; and gradual unfreezing, which unfreezes the parameters one layer at a time.

\paragraph{Masked LM and next-sentence prediction} BERT \cite{Devlin2018} combines both word and sentence representations (via masked LM and next sentence prediction objectives) in a single very large pretrained transformer \cite{Vaswani2017AttentionIA}.  It is adapted to both word and sentence level tasks by \FT~with task-specific layers.

%\paragraph{Masked LM and next-sentence prediction} BERT-base \cite{Devlin2018} uses a 12-layer transformer model, which is pretrained using a masked variant of the language modeling loss and a next-sentence prediction task similar to earlier work \cite{Kiros2015a,Logeswaran2018}. For adaptation, the model is fine-tuned with task-specific layers.

% \begin{itemize}
%     \item sentence embedding methods
%     \item ELMo 2-layer bidirectional language model, LSTM
%     \item BERT-base 12 layer transformer with next sentence prediction + masked language modeling
% \end{itemize}

\section{Experimental Setup}

%We compare ELMo and BERT as representatives of the two best-performing pretraining settings, optionally using ULMFiT techniques.
We compare ELMo and BERT as representatives of the two best-performing pretraining settings.
This section provides an overview of our methods; see the supplement  for full details.

\subsection{Target Tasks and Datasets}
%To cover a diverse range of target tasks, we evaluate on seven datasets and five tasks: named entity recognition (NER), sentiment analysis (SA), and three sentence pair tasks, natural language inference (NLI), paraphrase detection (PD), and semantic textual similarity (STS).
We evaluate on a diverse set of target tasks: named entity recognition (NER), sentiment analysis (SA), and three sentence pair tasks, natural language inference (NLI), paraphrase detection (PD), and semantic textual similarity (STS).

%We use the CoNLL 2003 \textbf{NER} dataset \citep{CoNLL2003NER}, which provides token level annotations of newswire across four different entity types (\texttt{PER}, \texttt{LOC}, \texttt{ORG}, \texttt{MISC}).
\paragraph{NER} We use the CoNLL 2003 dataset \citep{CoNLL2003NER}, which provides token level annotations of newswire across four different entity types (\texttt{PER}, \texttt{LOC}, \texttt{ORG}, \texttt{MISC}).

%\paragraph{SA} We use the binary version of the Stanford Sentiment Treebank \citep[SST-2;][]{socher2013recursive}, providing sentiment labels (\texttt{negative} or \texttt{positive}) of sentences from movie reviews. \nascomment{clarify:  full sentence classification only?}

\paragraph{SA} We use the binary version of the Stanford Sentiment Treebank \citep[SST-2;][]{socher2013recursive}, providing sentiment labels (\texttt{negative} or \texttt{positive}) for phrases and sentences of movie reviews.
%We use the binary version of the Stanford Sentiment Treebank \citep[SST-2;][]{socher2013recursive}, providing sentiment labels (\texttt{negative} or \texttt{positive}) for phrases and sentences of movie reviews.

%NLI is the task of determining whether a hypothesis is true, neutral, or contradictory given a premise.
\paragraph{NLI} We use both the broad-domain MultiNLI dataset \cite{Williams2017ABC} and Sentences Involving Compositional Knowledge \cite[SICK-E;][]{marelli2014sick}.
%For \textbf{NLI}, we use both the broad-domain MultiNLI dataset \cite{Williams2017ABC} and Sentences Involving Compositional Knowledge \cite[SICK-E;][]{marelli2014sick}.
%We use the dataset MultiNLI \cite{Williams2017ABC} due to its broad domain coverage. We additonally use Sentences Involving Compositional Knowledge \cite[SICK-E;][]{marelli2014sick}.
%(five different domains for training and evaluation, plus an additional five heldout domains for evaluation).

\paragraph{PD} For paraphrase detection (i.e., decide whether two sentences are semantically equivalent), we use the Microsoft Research Paraphrase Corpus \cite[MRPC;][]{dolan2005automatically}.
%For \textbf{PD} we use the Microsoft Research Paraphrase Corpus \cite[MRPC;][]{dolan2005automatically}.
%, a corpus of sentence pairs automatically extracted from news articles and then human annotated.

\paragraph{STS} We employ the Semantic Textual Similarity Benchmark \cite[STS-B;][]{Cer2017SemEval2017T1} and SICK-R \cite{marelli2014sick}. Both datasets, provide a human judged similarity value from 1 to 5 for each sentence pair.
%We employ two \textbf{STS} datasets, the Semantic Textual Similarity Benchmark \cite[STS-B;][]{Cer2017SemEval2017T1} and SICK-R \cite{marelli2014sick}. Both provide a human judged similarity value from 1 to 5 for each sentence pair.

\subsection{Adaptation}

We now describe how we adapt ELMo and BERT to these tasks. For \FE~we require a task-specific architecture, while for \FT~we need a task-specific output layer.
%In addition, the task-specific layers depend on the pretraining task and model.
% TODO: ADD DISCUSSON OF RELATIONSHIP BETWEEN TASK OBJECTIVE AND Pre-TRAIN OBJECTIVE
%For fair comparison, we conduct an extensive hyperparameter search over different learning rates and other hyperparameters for each task.
For fair comparison, we conduct an extensive hyper-parameter search for each task.

\paragraph{Feature extraction (\FE)} For both ELMo and BERT, we extract contextual representations of the words from all layers. During adaptation, we learn a linear weighted combination of the layers \cite{Peters2018} which is used as input to a task-specific model.
When extracting features, it is important to expose the internal layers as they typically encode the most transferable representations.
For SA, we employ a bi-attentive classification network \cite{Mccann2017}. For the sentence pair tasks, we use the ESIM model \cite{Chen2017}. For NER, we use a BiLSTM with a CRF layer \cite{CRF:Lafferty2001,Lample2016}.
% Then:
% \begin{itemize}
%     \item classification: use bi-attentive classification network
%     \item NLI / similarity: use ESIM
%     \item NER: use biLSTM + CRF
% \end{itemize}

\begin{table*}[]
\resizebox{\textwidth}{!}{%
\begin{tabular}{c c r r rr rrr}
\toprule
\multirow{2}{*}{\textbf{Pretraining}} & \multirow{2}{*}{\textbf{Adaptation}} & \textbf{NER} & \textbf{\textbf{SA}} & \multicolumn{2}{c}{\textbf{Nat. lang. inference}} & \multicolumn{3}{c}{\textbf{Semantic textual similarity}} \\
 &  & \textbf{CoNLL 2003} & \textbf{SST-2} & \textbf{MNLI} & \textbf{SICK-E} & \textbf{SICK-R} & \textbf{MRPC} & \textbf{STS-B}  \\ \midrule
%\multirow{2}{*}{\textbf{Pretraining}} & \multirow{2}{*}{\textbf{Adaptation}} & \textbf{SA} & \textbf{\textbf{NER}} & \multicolumn{2}{c}{\textbf{Nat. lang. inference}} & \multicolumn{3}{c}{\textbf{Semantic textual similarity}} \\
% &  & \textbf{SST-2} & \textbf{CoNLL 2003} & \textbf{MNLI} & \textbf{SICK-E} & \textbf{SICK-R} & \textbf{MRPC} & \textbf{STS-B}  \\ \midrule
Skip-thoughts & \FE & - & 81.8 & 62.9 & - & 86.6 & 75.8 & 71.8 \\
% InferSent & \FE & 84.6 & - & 66.1 & 86.3 & 88.4 & 76.2 & 75.9 \\ 
%Quick-thoughts & \FE & - & - & - & - & 87.4 & 76.9 & - \\
% BERT-large & \FT & 94.9 & 92.8 & 86.7 & - & 89.3 & 87.6 & - \\
\midrule
\multirow{3}{*}{ELMo} & \FE & 91.7 & \textbf{91.8} & \textbf{79.6} & \textbf{86.3} & \textbf{86.1} & \textbf{76.0} & \textbf{75.9} \\
 & \FT & \textbf{91.9} & 91.2 & 76.4 & 83.3 & 83.3 & 74.7 & 75.5 \\
& $\Delta$=\FT-\FE & 0.2 & -0.6 & \cellcolor{fe_blue}-3.2 & \cellcolor{fe_blue}-3.3 & \cellcolor{fe_blue}-2.8 & \cellcolor{fe_blue}-1.3 & -0.4 \\
%\multirow{3}{*}{ELMo} & \FE & \textbf{91.8} & 91.7 & \textbf{79.6} & \textbf{86.3} & \textbf{86.1} & \textbf{76.0} & \textbf{75.9} \\
% & \FT & 91.2 & \textbf{91.9} & 76.4 & 83.3 & 83.3 & 74.7 & 75.5 \\
%& $\Delta$=\FT-\FE & -0.6 & 0.2 & \cellcolor{fe_blue}-3.2 & \cellcolor{fe_blue}-3.3 & \cellcolor{fe_blue}-2.8 & \cellcolor{fe_blue}-1.3 & -0.4 \\
 \midrule
 \multirow{3}{*}{BERT-base} & \FE & 92.2 & 93.0 & \textbf{84.6} & 84.8 & 86.4 & 78.1 & 82.9 \\
 & \FT & \textbf{92.4} & \textbf{93.5} & \textbf{84.6} & \textbf{85.8} & \textbf{88.7} & \textbf{84.8} & \textbf{87.1} \\
& $\Delta$=\FT-\FE & 0.2 & 0.5 & 0.0 & 1.0 & \cellcolor{fe_red}2.3 & \cellcolor{fe_red}6.7 & \cellcolor{fe_red}4.2 \\
%\multirow{3}{*}{BERT-base} & \FE & 93.0 & 92.2 & \textbf{84.6} & 84.8 & 86.4 & 78.1 & 82.9 \\
% & \FT & \textbf{93.5} & \textbf{92.4} & \textbf{84.6} & \textbf{85.8} & \textbf{88.7} & \textbf{84.8} & \textbf{87.1} \\
%& $\Delta$ = \FT-\FE & 0.5 & 0.2 & 0.0 & 1.0 & \cellcolor{fe_red}2.3 & \cellcolor{fe_red}6.7 & \cellcolor{fe_red}4.2 \\
\bottomrule
\end{tabular}
}
\caption{Test set performance of feature extraction (\FE) and fine-tuning (\FT) approaches for ELMo and BERT-base compared to two sentence embedding methods.
%For \FT, we only show the difference vs. \FE~for clarity.
Settings that are good for \FT~are colored in \textcolor{fe_red}{red} ($\Delta$=\FT-\FE~$>$ 1.0); settings good for \FE~are colored in \textcolor{fe_blue}{blue} ($\Delta$=\FT-\FE~$<$ -1.0).
Numbers for baseline methods are from respective papers, except for SST-2, MNLI, and STS-B results, which are from \citet{Wang2018a}. BERT fine-tuning results (except on SICK) are from \citet{Devlin2018}.
The metric varies across tasks (higher is always better): accuracy for SST-2, SICK-E, and MRPC; matched accuracy for MultiNLI; Pearson correlation for STS-B and SICK-R; and span F$_1$ for CoNLL 2003.
%For each task and model, the best performing approach is underlined.
For CoNLL 2003, we report the mean with five seeds; standard deviation is about 0.2\%.}
\label{tab:ff_ft_elmo_bert}

\end{table*}

\paragraph{Fine-tuning (\FT): ELMo} We max-pool over the LM states and add a softmax layer for text classification. For the sentence pair tasks, we compute cross-sentence bi-attention between the LM states \cite{Chen2017}, apply a pooling operation, then add a softmax layer. For NER, we add a CRF layer on top of the LSTM states.

\paragraph{Fine-tuning (\FT): BERT} We feed the sentence representation into a softmax layer for text classification and sentence pair tasks following \citet{Devlin2018}.
%For sentence pair tasks, we initially concatenate the sentences with a \texttt{[SEP]} token.
For NER, we extract the representation of the first word piece for each token and add a softmax layer.

\section{Results}

% Quick-thoughts, on semantic similarity, paraphrasing haven't gotten look, InferSent 

%We show results in Table \ref{tab:ff_ft_elmo_bert} comparing ELMo and BERT for both \FE~and \FT~approaches across the seven tasks with two sentence embedding methods, Skip-thoughts \cite{Kiros2015a} and Quick-thoughts \cite{Logeswaran2018}, which employ a next-sentence prediction objective similar to BERT.
We show results in Table \ref{tab:ff_ft_elmo_bert} comparing ELMo and BERT for both \FE~and \FT~approaches across the seven tasks with one sentence embedding method, Skip-thoughts \cite{Kiros2015a}, that employs a next-sentence prediction objective similar to BERT.

%The overall performance of feature extraction and fine-tuning is about the same for both pretrained models and for most tasks, except for the paraphrase and sentence prediction tasks.

Both ELMo and BERT outperform the sentence embedding method significantly, except on the semantic textual similarity tasks (STS) where Skip-thoughts is similar to ELMo.
%BERT mostly outperforms ELMo, particularly on the STS tasks.
The overall performance of \FE~and \FT~varies from task to task, with small differences except for a few notable cases.
% Accordingly, we recommend in most cases practitioners choose an approach based on practical considerations, e.g., computational expense and ease of integration with existing architectures.
For ELMo, we find the largest differences for sentence pair tasks where \FE~consistently outperforms~\FT.
%In many cases for the ELMo LSTM LM, fine-tuning required large hyperparameter searches, potentially due to difficulties fine-tuning LSTMs (see discussion below).
For BERT, we obtain nearly the opposite result: \FT~significantly outperforms \FE~on all STS tasks, with much smaller differences for the others.
%This is particularly striking as SICK-E and SICK-R labels are fairly well correlated \cite{marelli2014sick}.

%For BERT, we observe the largest differences in the performance of feature extraction and fine-tuning on all three datasets of the two semantic similarity tasks, paraphrase detection and semantic textual similarity---with marginal differences for the others. This is particularly striking as SICK-E and SICK-R labels are fairly well correlated \cite{marelli2014sick}.

\paragraph{Discussion} Past work in NLP \cite{Mou2016} showed that similar pretraining tasks transfer better.\footnote{\citet{Mou2016}, however, only investigate transfer between classification tasks (NLI $\rightarrow$ SICK-E/MRPC).} In computer vision (CV), \citet{Yosinski2014HowTA} similarly found that the transferability of features decreases as the distance between the pretraining and target task increases.
%In this vein, Skip-thoughts and Quick-thoughts, which use a next-sentence prediction objective similar to BERT, perform particularly well on semantic textual similarity (STS) tasks, which indicates a close alignment between the pretraining and target task.
In this vein, Skip-thoughts---and Quick-thoughts \cite{Logeswaran2018}, which has similar performance---which use a next-sentence prediction objective similar to BERT, perform particularly well on STS tasks, indicating a close alignment between the pretraining and target task.
This strong alignment also seems to be the reason for BERT's strong relative performance on these tasks. 

In CV, \FT~generally outperforms \FE~when transferring from ImageNet supervised classification pretraining to other classification tasks \cite{Kornblith2018}. Recent results suggest \FT~is less useful for more distant target tasks such as semantic segmentation \cite{He2018}. This is in line with our results, which show strong performance with \FT~between closely aligned tasks (next-sentence prediction in BERT and STS tasks) and poor performance for more distant tasks (LM in ELMo and sentence pair tasks). A confounding factor may be the suitability of the inductive bias of the model architecture for sentence pair tasks, which we will analyze next.

\section{Analyses}

\paragraph{Modelling pairwise interactions} LSTMs consider each token sequentially, while Transformers can relate each token to every other in each layer \cite{Vaswani2017AttentionIA}. This might facilitate \FT~with Transformers on sentence pair tasks, on which ELMo-\FT~performs comparatively poorly. To analyze this further, we compare different ways of encoding the sentence pair with ELMo and BERT. For ELMo, we compare encoding with and without cross-sentence bi-attention in Table \ref{tab:elmo-cross-sentence}.
When adapting the ELMo LSTM to a sentence pair task, modeling the sentence interactions by fine-tuning through the bi-attention mechanism provides the best performance.\footnote{This is similar to text classification tasks, where we find max-pooling to outperform using the final hidden state, similar to \cite{Howard2018}.} This provides further evidence that the LSTM has difficulty modeling the pairwise interactions during sequential processing.
This is in contrast to a Transformer LM that can be fine-tuned in this manner \cite{Radford2018}.

\label{sec:pairwise_analysis}

\begin{table}[]
\resizebox{\columnwidth}{!}
{%
%\begin{tabular}{l p{7ex} p{7ex} p{7ex} p{6ex}}
\begin{tabular}{l c c c c}
\toprule
 & \textsc{SICK-E} & \textsc{SICK-R} & \textsc{STS-B} & \textsc{MRPC} \\
 \midrule
ELMo-\FT~+bi-attn. & 83.8 & 84.0 & 80.2 & 77.0 \\
$\>$ w/o bi-attn.  & 70.9 & 51.8 & 38.5 & 72.3 \\
\bottomrule
\end{tabular}%
}
\caption{Comparison of ELMO-\FT~ cross-sentence embedding methods on dev.~sets of sentence pair tasks.}
\label{tab:elmo-cross-sentence}
\end{table}

%We find that when adapting a pretrained LSTM to a sentence pair task, the LSTM is not able to model the pairwise interactions between sentences if the sentences are only processed sequentially as in the transformer. We need to add a self-attention layer to achieve good performance on such tasks. This is similar to text classification tasks, where we find max-pooling to outperform using the final hidden state, similar to \cite{Howard2018}.

% We hypothesize that BERT's strong performance pairwise tasks is due to the transformer's ability to encode cross-sentence relations.
For BERT-\FE, we compare joint encoding of the sentence pair with encoding the sentences separately in Table \ref{tab:bert-cross-sentence}. The latter leads to a drop in performance, which shows that the BERT representations encode cross-sentence relationships and are therefore particularly well-suited for sentence pair tasks.

\begin{table}[]
\resizebox{\columnwidth}{!}
{%
%\begin{tabular}{l p{7ex} p{7ex} p{7ex} p{6ex}}
\begin{tabular}{l c c c c}
\toprule
 & \textsc{SICK-E} & \textsc{SICK-R} & \textsc{STS-B} & \textsc{MRPC} \\
 \midrule
BERT-\FE, joint enc.  & \textbf{85.5} & 86.4 & \textbf{88.1} & \textbf{83.3} \\ 
$\>$ separate encoding & 81.2 & \textbf{86.8} & 86.8 & 81.4 \\
\bottomrule
\end{tabular}%
}
\caption{Comparison of BERT-\FE~cross-sentence embedding methods on dev.~sets of sentence pair tasks.}
\label{tab:bert-cross-sentence}
\end{table}

% Ablations:
% \begin{itemize}
%     \item when tuning LSTM architecutures (ELMo) that need to compress all information in a sequence into a vector, approaches that tune contextual word representations do best (pooling over sequence vs using final state for classification and for NLI fine tuning through attention works better then linearizing sentence pair ala open ai).
%     \item for bert, in sentence pair tasks extracting features from both sentences together works better then extracting features from each sentence individually.  Shows that bert encodes cross sentence relationships in its word embeddings and is therefore particularly well suited for tuning sentence pair classification tasks
% \end{itemize}

\paragraph{Impact of additional parameters} We evaluate whether adding  parameters is useful for both adaptation settings on NER. We add a CRF layer (as used in \FT) and a BiLSTM with a CRF layer (as used in \FE) to both and show results in Table \ref{tab:ner-ablations}. We find that additional parameters are key for \FE, but hurt performance with \FT. In addition, \FT~requires gradual unfreezing \cite{Howard2018} to match performance of feature extraction.

\paragraph{ELMo fine-tuning} We found fine-tuning the ELMo LSTM to be initially difficult and required careful hyper-parameter tuning.  Once tuned for one task, other tasks have similar hyper-parameters.  Our best models used slanted triangular learning rates and discriminative fine-tuning \cite{Howard2018} and in some cases gradual unfreezing.
%Fine-tuning the transformer required a similar learning rate schedule, but was otherwise less sensitive to hyperparameter choices.

%%%%%%%%%%%%%% TABLE OF ABLATIONS FOR NER
\begin{table}[]
\centering
%\resizebox{\columnwidth}{!}{%
\begin{tabular}{l r}
\toprule
\textbf{Model configuration} & \textbf{F$_1$} \\
 \midrule
 \FE~+ BiLSTM + CRF   & \textbf{95.5} \\
 \FE~+ CRF & 91.9 \\
 \midrule
 \FT~+ CRF + gradual unfreeze & \textbf{95.5} \\
 \FT~+ BiLSTM + CRF + gradual unfreeze& 95.2 \\
 \FT~+ CRF & 95.1 \\
\bottomrule
\end{tabular}%
%}
\caption{Comparison of CoNLL 2003 NER development set performance (F$_1$) for ELMo for both feature extraction and fine-tuning.  All results averaged over five random seeds.
}
\label{tab:ner-ablations}
\end{table}

\paragraph{Impact of target domain} Pretrained language model representations are intended to be universal.
However, the target domain might still impact the adaptation performance.
We calculate the Jensen-Shannon divergence based on term distributions \citep{Ruder2017} between the domains used to train BERT (books and Wikipedia) and each MNLI domain.
%We calculate the Jensen-Shannon similarity based on term distributions \citep{Ruder2017} between the domains used to train BERT (books and Wikipedia) and each MNLI domain.\footnote{We define the similarity $sim = 1 - JS$ where $JS$ is the Jensen-Shannon divergence. We computed similar values for the $\mathcal{A}$ distance \citep{Blitzer2007}.}
%We show results in Table \ref{tab:bert-elmo-domain-impact}. We find no significant correlation. At least for this task \nascomment{and this way of quantifying domain similarity?}, the \nascomment{distance of the} target domain does not seem to have a major impact on the adaptation performance.
We show results in Table \ref{tab:bert-elmo-domain-impact}. We find no significant correlation. At least for this task, the distance of the source and target domains does not seem to have a major impact on the adaptation performance.

\begin{table}[]
\resizebox{\columnwidth}{!}{%
\begin{tabular}{l c c c c c}
\toprule
%  & \textsc{te} & \textsc{fi} & \textsc{tr} & \textsc{go} & \textsc{sl} \\ % & All \\
  & \textsc{te} & \textsc{go} & \textsc{tr} & \textsc{fi} &  \textsc{sl} \\ % & All \\
 \midrule
% \# train examples & 83,348 & 77,348 & 77,350 & 77,350 & 77,306 & 392,702 \\
%BERT-\FT & 83.32 & 84.86 & 85.48 & 86.52 & 80.28 & 84.10\\
BERT-\FE & 84.4 & 86.7 & 86.1 & 84.5 & 80.9 \\ % & 84.5 \\
% BERT-\FT & 83.3 & 84.9 & 85.5 & 86.5 & 80.3 \\ % & 84.1 \\
$\Delta$=\FT-\FE & -1.1 & -0.2 &  -0.6 & 0.4 & -0.6 \\ % -0.4 \\
% $\mathcal{A}$ dist BERT & \\
%JS sim & .794 & .910 & .863 & .820 & .914 \\ % \\ & .929 \\
JS div & 0.21 & 0.18 & 0.14 & 0.09 & 0.09 \\ % \\ & .929 \\
% \midrule
% ELMo & \\
% $\mathcal{A}$ dist ELMo & \\
% JS sim ELMo & .774 & .854 & .863 & .866 & .932 \\ % & .926\\
% Perplexity & & \\
\bottomrule
\end{tabular}%
}
\caption{Accuracy of feature extraction (\FE) and fine-tuning (\FT) with BERT-base trained on training data of different MNLI domains and evaluated on corresponding dev sets. \textsc{te}: telephone. \textsc{fi}: fiction. \textsc{tr}: travel. \textsc{go}: government. \textsc{sl}: slate.}
\label{tab:bert-elmo-domain-impact}
\end{table}

\paragraph*{Representations at different layers} In addition, we are interested how the information in the different layers of the models develops over the course of fine-tuning. We measure this information in two ways: a) with diagnostic classifiers \citep{Adi2017}; and b) with mutual information \citep[MI;][]{Noshad2018}. Both methods allow us to associate the hidden activations of our model with a linguistic property. In both cases, we use the mean of the hidden activations of BERT-base\footnote{We show results for BERT as they are more inspectable due to the model having more layers. Trends for ELMo are similar.} of each token / word piece of the sequence(s) as the representation.\footnote{We observed similar results when using max-pooling or the representation of the first token.}

\begin{figure}[h]
\centering
\includegraphics[width=0.9\linewidth]{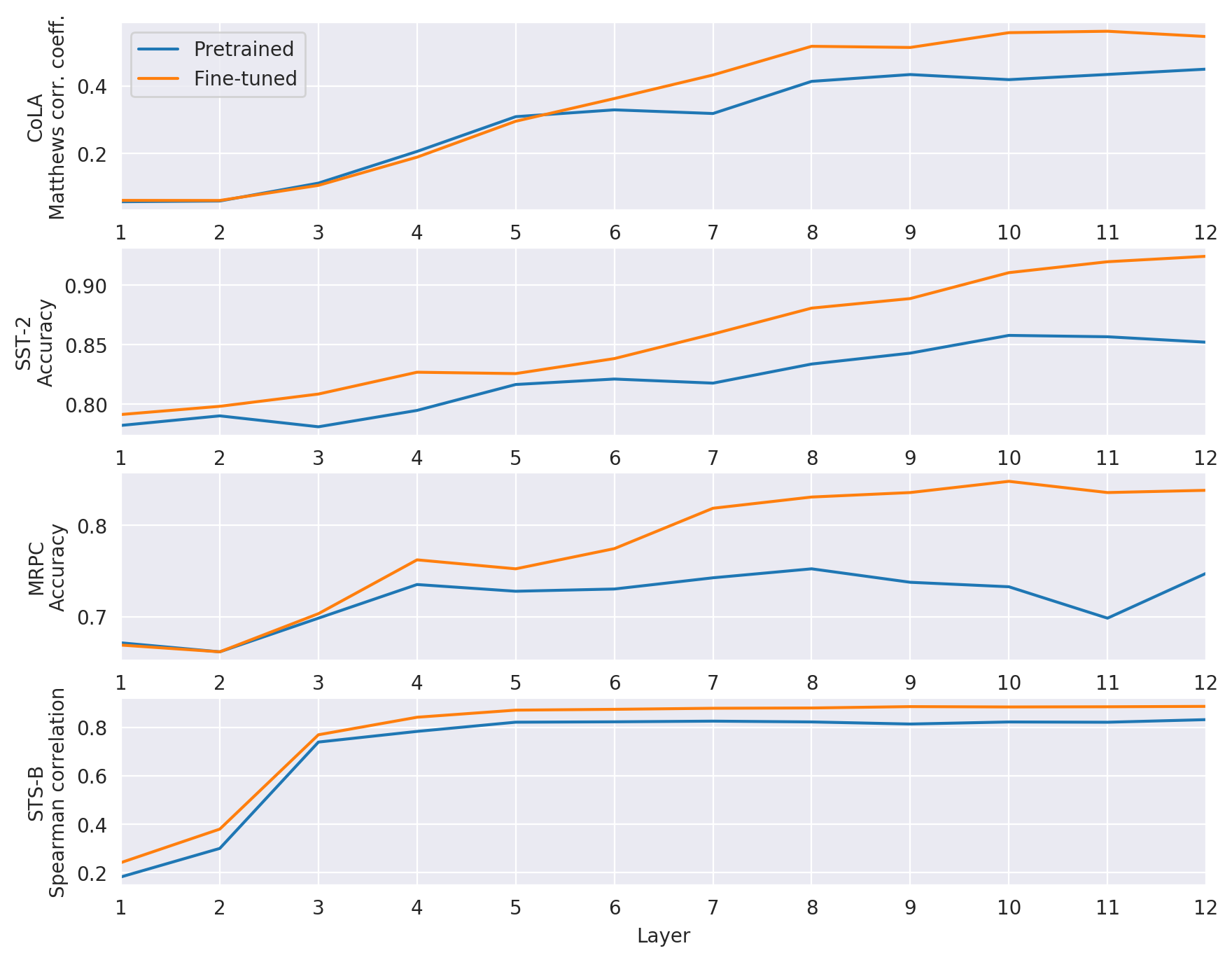}
\caption[Performance of diagnostic classifiers trained on BERT representations]{Performance of diagnostic classifiers trained on pretrained and fine-tuned BERT representations at different layers on the dev sets of the corresponding tasks.}
\label{fig:diagnostic_classifiers}
\end{figure}

With diagnostic classifiers, for each example, we extract the pretrained and fine-tuned representation at each layer as features. We use these features as input to train a logistic regression model (linear regression for STS-B, which has real-valued outputs) on the training data of two single sentence (CoLA\footnote{The Corpus of Linguistic Acceptability (CoLA) consists of examples of expert English sentence acceptability judgments drawn from 22 books and journal articles on linguistic theory. It uses the Matthews correlation coefficient \citep{matthews1975comparison} for evaluation and is available at: \url{nyu-mll.github.io/CoLA}} and SST-2) and two pair sentence tasks (MRPC and STS-B). We show its performance on the corresponding dev sets in Figure \ref{fig:diagnostic_classifiers}.

For all tasks, diagnostic classifier performance generally is higher in higher layers of the model. Fine-tuning improves the performance of the diagnostic classifier at every layer. For the single sentence classification tasks CoLA and SST-2, pretrained performance increases gradually until the last layers. In contrast, for the sentence pair tasks MRPC and STS-B performance is mostly flat after the fourth layer. Relevant information for sentence pair tasks thus does not seem to be concentrated primarily in the upper layers of pretrained representations, which could explain why fine-tuning is particularly useful in these scenarios.

Computing the mutual information with regard to representations of deep neural networks has only become feasible recently with the development of more sophisticated MI estimators. In our experiments, we use the state-of-the-art ensemble dependency graph estimator \citep[EDGE;][]{Noshad2018} with default hyper-parameter values. As a sanity check, we compute the MI between hidden activations and random labels and random representations and random labels, which yields $0$ in every case as we would expect.\footnote{For the same settings, we obtain non-zero values with earlier estimators \citep{Saxe2018}, which seem to be less reliable for higher numbers of dimensions.}

\begin{figure}[h]
\centering
\includegraphics[width=0.9\linewidth]{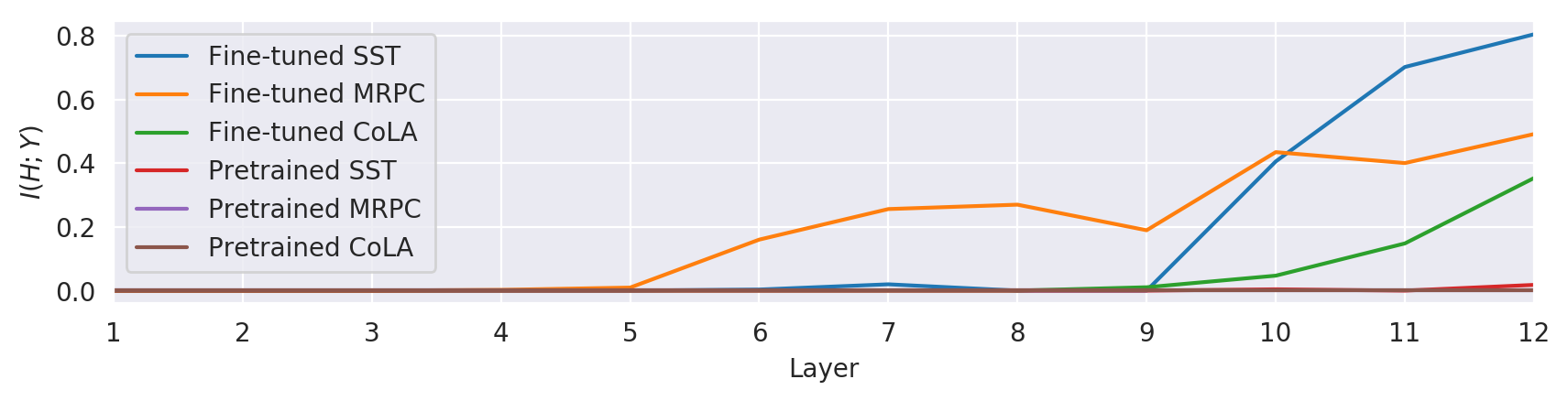}
\caption[Mutual information between BERT representations and labels]{The mutual information between fine-tuned and pretrained mean BERT representations and the labels on the dev set of the corresponding tasks.}
\label{fig:mutual_information}
\end{figure}

We show the mutual information $I(H; Y)$ between the pretrained and fine-tuned mean hidden activations $H$ at each layer of BERT and the output labels $Y$ on the dev sets of CoLA, SST-2, and MRPC in Figure \ref{fig:mutual_information}.

The MI between pretrained representations and labels is close to $0$ across all tasks and layers, except for SST where the last layer shows a small non-zero value. In contrast, fine-tuned representations display much higher MI values. The MI for fine-tuned representations rises gradually through the intermediate and last layers for the sentence pair task MRPC, while for the single sentence classification tasks, the MI rises sharply in the last layers. Similar to our findings with diagnostic classifiers, knowledge for single sentence classification tasks thus seems mostly concentrated in the last layers, while pair sentence classification tasks gradually build up information in the intermediate and last layers of the model.

\section{Conclusion}

We have empirically analyzed fine-tuning and feature extraction approaches across diverse datasets, finding that the relative performance depends on the similarity of the pretraining and target tasks. We have explored possible explanations and provided practical recommendations for adapting pretrained representations to NLP practicioners.

\bibliography{naaclhlt2019}
\bibliographystyle{acl_natbib}

\appendix

\section{Experimental details}

For fair comparison, all experiments include extensive hyper-parameter tuning.
We tuned the learning rate, dropout ratio, weight decay and number of training epochs.
In addition, the fine-tuning experiments also examined the impact of triangular learning rate schedules, gradual unfreezing, and discriminative learning rates.
Hyper-parameters were tuned on the development sets and the best setting evaluated on the test sets.

All models were optimized with the Adam optimizer \cite{Kingma2014AdamAM} with weight decay fix \cite{Loshchilov2017FixingWD}. 

We used the publicly available pretrained ELMo\footnote{https://allennlp.org/elmo} and BERT\footnote{https://github.com/google-research/bert} models in all experiments.
For ELMo, we used the original two layer bidirectional LM.
In the case of BERT, we used the BERT-base model, a 12 layer bidirectional transformer.  We used the English uncased model for all tasks except for NER which used the English cased model.

\subsection{Feature extraction}

To isolate the effects of fine-tuning contextual word representations, all feature based models only include one type of word representation (ELMo or BERT) and do not include any other pretrained word representations.

For all tasks, all layers of pretrained representations were weighted together with learned scalar parameters following \citet{Peters2018}.

\paragraph{NER} For the NER task, we use a two layer bidirectional LSTM in all experiments.
For ELMo, the output layer is a CRF, similar to a state-of-the-art NER system \cite{Lample2016}.  Feature extraction for ELMo treated each sentence independently.

In the case of BERT, the output layer is a softmax to be consistent with the fine-tuned experiments presented in \citet{Devlin2018}.  In addition, as in \citet{Devlin2018}, we used document context to extract word piece representations.
When composing multiple word pieces into a single word representation, we found it beneficial to run the biLSTM layers over all word pieces before taking the LSTM states of the first word piece in each word.  We experimented with other pooling operations to combine word pieces into a single word representation but they did not provide additional gains.

\paragraph{SA}  We used the implementation of the bi-attentive classification network in AllenNLP \cite{Gardner2017AllenNLP} with default hyper-parameters, except for tuning those noted above.  As in the fine-tuning experiments for SST-2, we used all available annotations during training, including those of sub-trees.  Evaluation on the development and test sets used full sentences.

\paragraph{Sentence pair tasks} When extracting features from ELMo, each sentence was handled separately.  For BERT, we extracted features for both sentences jointly to be consistent with the pretraining procedure.  As reported in Section 5 this improved performance over extracting features for each sentence separately.

Our model is the ESIM model \cite{Chen2017}, modified as needed to support regression tasks in addition to classification.  We used default hyper-parameters except for those described above.

\subsection{Fine-tuning}

When fine-tuning ELMo, we found it beneficial to use discriminative learning rates \cite{Howard2018} where the learning rate decreased by $0.4 \times$ in each layer (so that the learning rate for the second to last layer is $0.4\times$ the learning rate in the top layer).
In addition, for SST-2 and NER, we also found it beneficial to gradually unfreeze the weights starting with the top layer. In this setting, in each epoch one additional layer of weights is unfrozen until all weights are training.  These settings were chosen by tuning development set performance.

For fine-tuning BERT, we used the default learning rate schedule \cite{Devlin2018} that is similar to the schedule used by \citet{Howard2018}.

\paragraph{SA} We considered several pooling operations for composing the ELMo LSTM states into a vector for prediction including max pooling, average pooling and taking the first/last states.  Max pooling performed slightly better than average pooling on the development set.

\paragraph{Sentence pair tasks} Our bi-attentive fine-tuning mechanism is similar to the the attention mechanism in the feature based ESIM model.  To apply it, we first computed the bi-attention between all words in both sentences, then applied the same ``enhanced'' pooling operation as in \cite{Chen2017} before predicting with a softmax.  Note that this attention mechanism and pooling operation does not add any additional parameters to the network.

\end{document}